\documentclass[10pt,twocolumn,letterpaper]{article}

\usepackage{cvpr}              

\usepackage{graphicx}
\usepackage{amsmath}
\usepackage{amssymb}
\usepackage{booktabs}
\usepackage{multirow}
\usepackage{tabularx}
\usepackage{caption}
\usepackage[section]{placeins}
\usepackage{placeins}
\usepackage{paralist}

\usepackage[pagebackref,breaklinks,colorlinks]{hyperref}

\usepackage[capitalize]{cleveref}
\crefname{section}{Sec.}{Secs.}
\Crefname{section}{Section}{Sections}
\Crefname{table}{Table}{Tables}
\crefname{table}{Tab.}{Tabs.}

\def\cvprPaperID{*****} 
\def\confName{CVPR}
\def\confYear{2022}

\begin{document}

\title{2CET-GAN: Pixel-Level GAN Model for Human Facial Expression Transfer}

\author{Xiaohang Hu\thanks{corresponding author}\\
{\tt\small xiaohanghu123@gmail.com}
\and
Nuha Aldausari\\
{\tt\small n.aldausari@unsw.edu.au}
\and
Gelareh Mohammadi\\
{\tt\small g.mohammadi@unsw.edu.au}
\and
University of New South Wales\\
}

\twocolumn[{
\maketitle
    \centering
    \captionsetup{type=figure}
    \setlength{\abovecaptionskip}{3pt} 
    \includegraphics[width=1.0\textwidth]{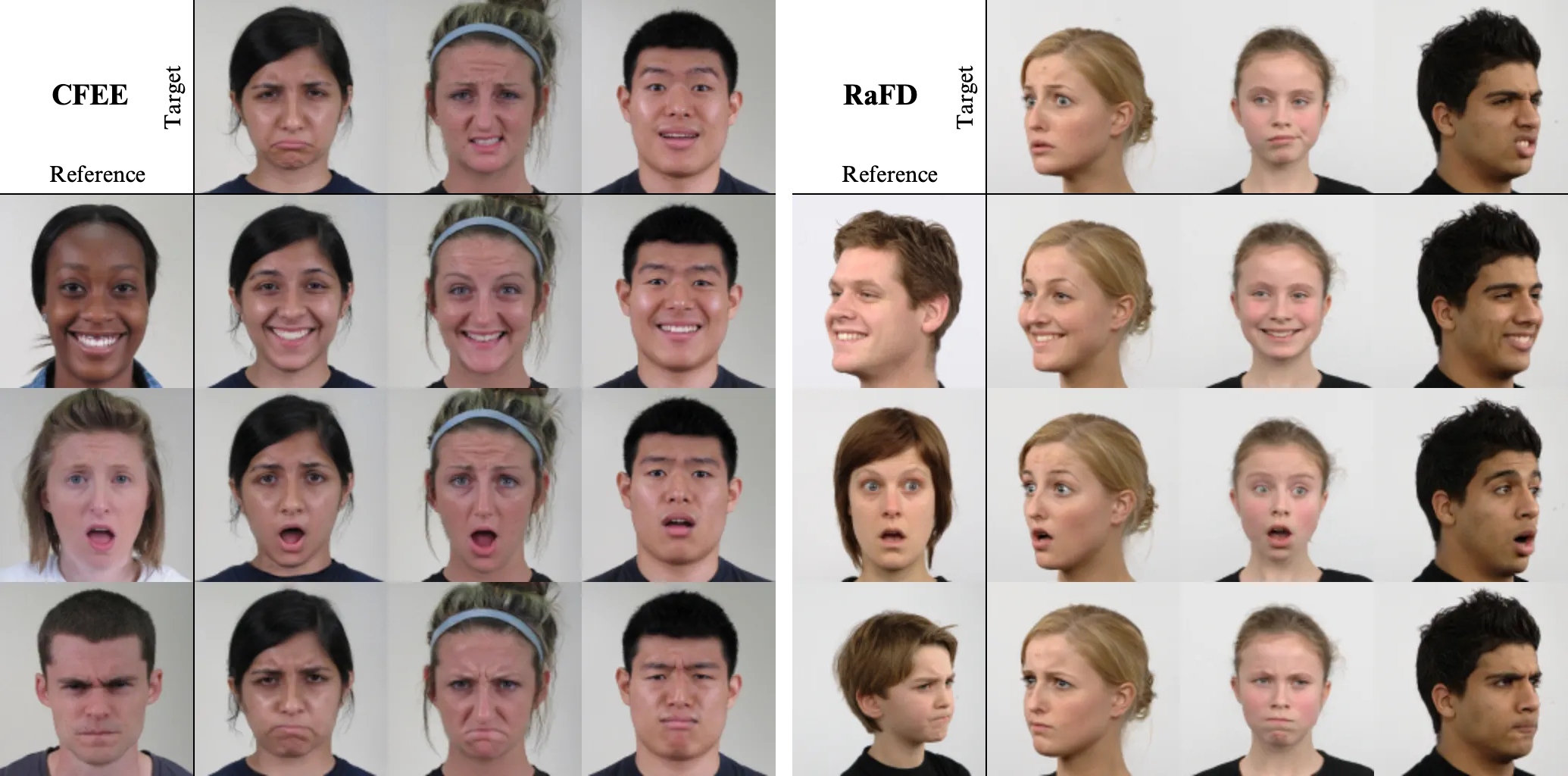}
    \begin{tabularx}{\textwidth}{>{\centering\arraybackslash}X >{\centering\arraybackslash}X}
        (a) Expression transfer on CFEE dataset & (b) Expression transfer on RaFD dataset\\
    \end{tabularx}
    \captionof{figure}{Expression transfer on the CFEE (a) and RafD (b) datasets by our approach. In both (a) and (b), the first row is target faces while the first column is reference faces. The remaining images are generated by our network (2CET-GAN). The synthesized faces took the identity from the target image and the expression from the corresponding reference image.}
    \label{fig:demo_matrix_1}
\begin{center}
\end{center}
}]

\begin{abstract}
Recent studies have used GAN to transfer expressions between human faces. However, existing models have many flaws: relying on emotion labels, lacking continuous expressions, and failing to capture the expression details. To address these limitations, we propose a novel CycleGAN- and InfoGAN-based network called 2 Cycles Expression Transfer GAN (2CET-GAN), which can learn continuous expression transfer without using emotion labels. The experiment shows our network can generate diverse and high-quality expressions and can generalize to unknown identities. To the best of our knowledge, we are among the first to successfully use an unsupervised approach to disentangle expression representation from identities at the pixel level. Our code is available at * (the code will be open-sourced after acceptance).
\end{abstract}

\section{Introduction}
\label{sec:intro}
Human facial expression transfer aims to transfer the expression from a reference face to a target face while maintaining the target's identity, which has exciting applications such as filmmaking, photo editing and entertainment. In \cref{fig:demo_matrix_1}, we show some examples of two emotion datasets and what an emotion transfer looks like. For example, the face located in the second row and second column of \cref{fig:demo_matrix_1}(a) takes the expression from the left reference face and the identity from the upper target face. This process is also known as face reenactment, according to some studies on videos \cite{garrido2014automatic}. The existing studies of human facial expression transfer can be divided into three categories: Three dimensional (3D) model-based methods, landmark/points-based methods, and pixel-based methods.


3D model based methods \cite{vlasic2006face,thies2016face2face,kim2018deep,garrido2016reconstruction,thies2018headon} assume the 3D shape of the face contains the expression. They first extract 3D face models from the target and reference faces receptively and then transfer the expression from the reference 3D model to the target 3D model, and finally combine the deformed target 3D model with the surface of the original face to generate the output. These steps can be implemented by traditional mathematical methods or can be replaced by deep learning models \cite{kim2018deep}. The 3D-based methods are mature and have been successfully used in movie and entertainment applications \cite{alexander2009creating}. However, these methods are complex and require enormous effort to assemble and tune multiple components.

To simplify the 3D-based methods, landmark-based methods \cite{qiao2018geometry,song2018geometry,zhang2020freenet,geng2018warp} were introduced that use fiducial points to represent the facial geometry. Similar to 3D-based methods, expression can be transferred from the reference landmarks to the target landmarks, but the landmarks can not directly blend with the face's surface since the landmarks do not contain sufficient geometric information. Thus, those methods usually employ a neural network to generate the final output by inputting the target image and the deformed target landmarks. Landmark-based generative models need large-scale datasets, and the intensity/number of points can limit the details of the generated expressions.

 Pixel-based methods \cite{pumarola2018ganimation,ding2018exprgan,ning2020emotiongan} directly transfer expressions at the pixel level. These methods usually employ neural networks to handle the complexities of extraction, transformation, and generation. The removal of the intermediate facial geometry means the model can potentially capture all expression details, such as tears and the color of the face. Some pixel-based models \cite{wang2021expression,choi2018stargan} take the discrete emotion labels as a conditional expression code, such as happy, sad and angry; and the training can be implemented by Conditional Generative Adversarial Network (CGAN) \cite{mirza2014conditional} in a supervised manner. The introduction of Cycle Consistency Loss (CycleGAN) \cite{zhu2017unpaired} removes the requirement of paired data for identity-preserving, which means each input does not need a corresponding true output for training. A further improvement was introduced in \cite{ding2018exprgan,ning2020emotiongan}, which through a “Controller Module”, maps discrete labels to a continuous space for generating continuous expressions.
However, almost all existing pixel-based methods directly or indirectly rely on expression labels \cite{ding2018exprgan}, and they hardly generate high-quality continuous expressions.

We have proposed a novel pixel-level GAN model that can transfer continuous expressions from one face to another without needing 3D/2D annotations or pre-trained models. We use only one label to distinguish natural faces from other expressions, unlike most models that use multiple expression labels. Our network contains an encoder and a generator. The encoder extracts diverse and continuous expression code from the reference face while the generator can apply the expression to the target face. Our contribution can be summed up as:
\begin{itemize}
  \item We propose a novel two cycles architecture that can learn facial expressions in an unsupervised manner. It only requires labeling neutral faces in a dataset.
  \item The proposed model can extract and transfer continuous expressions, which can capture more details and provides flexibility to applications.
  \item It provides a GAN-based architecture that can learn the transfer between any two distributions while preserving identity information.
  This architecture can be further applied to learn the expression transfer on 3D shapes or landmarks.
  \item We provide qualitative and quantitative evaluations on two human emotion datasets and show our model is superior compared to baseline models.
\end{itemize}

\section{Related work}
\label{sec:Related works}
In this section, we briefly discuss the related work that we have used or got inspiration from in the proposed model.

\textbf{GAN.} Generative Adversarial Network (GAN) \cite{goodfellow2020generative} has been widely used in human face synthesis. A generator and a discriminator are the two main components of GANs. In the face synthesis paradigm, a generator synthesizes a human face and the discriminator judges whether the given sample is from the real distribution. The generator competes with the discriminator through the adversarial loss, and they learn from the error in each iteration in an unsupervised manner. 

\textbf{CycleGAN.} Cycle-Consistent Adversarial Networks (CycleGAN)\cite{zhu2017unpaired} solves  the identity-preserving problem by introducing cycle consistency loss. It adds an inverse translator $F$ to transfer the generated fake face $G(x)$ back to the original face $x$, and computes the $L_1$ distance between the original face and generated-back face as the cycle consistency loss: $\left \| x-F(G(x)) \right \|_{1}$. Thus, the forward generator in CycleGAN keeps the identity information.

\begin{figure*}[!pt]
  \centering
  \includegraphics[width=\textwidth]{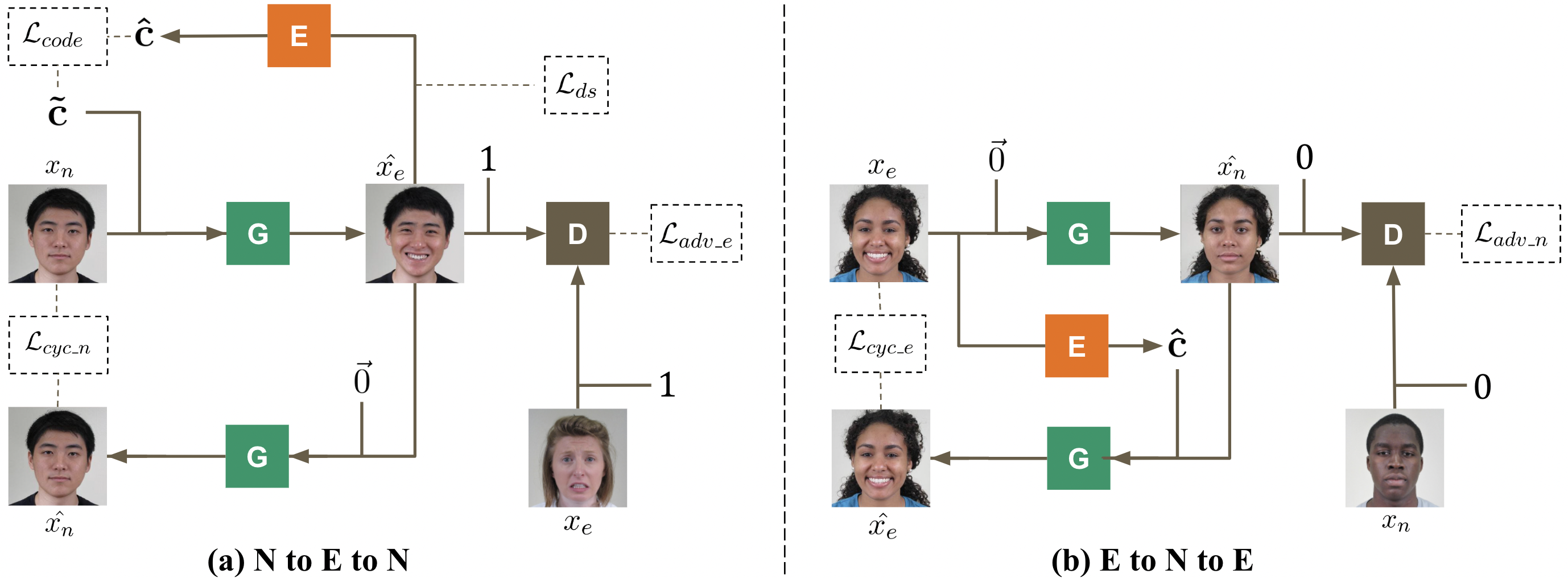}
  \caption{\textbf{The structure of 2CET-GAN} contains 2 cycles: (a) Neutral face to Emotional face to Neutral face (N to E to N); (b) Emotional face to Neutral face to Emotional face (E to N to E). The meaning of each colored block is: E represents the encoder; G represents the generator; D represents the discriminator. $\mathbf{\tilde{c}}$ denote randomly sampled expression code, $\mathbf{\hat{c}}$ denote extracted expression code. Dotted blocks represent loss functions.}
  \label{fig:structure}
\end{figure*}

\textbf{InfoGAN.} Information Maximizing Generative Adversarial Networks (InfoGAN) \cite{chen2016infogan} learns disentangled representations in an unsupervised manner by maximizing the mutual information between a random latent code and the observations. The generator $G$ takes a random latent code $\tilde{c}$ as an additional input, the discriminator $D$ extracts the code out from the fake image, and then the "info loss" $\left \| \tilde{c}-D(G(z, \tilde{c})) \right \|_{1}$ is used to maximize the mutual information. 
Thus, the latent code $\tilde{c}$ will correspond to some semantic features of the data, such as the pose, color or expression of the face. We employ a continuous multidimensional latent code as the expression representation and successfully disentangle the representation over diverse identities.

\textbf{StarGAN v2} \cite{choi2020starganv2} also adopts a CycleGAN- and InfoGAN-based architecture and it utilizes the diversity sensitive loss \cite{mao2019mode,yang2019diversity} to push the generator to explore the instance space more widely to generate diverse styles.
The experiment of StarGAN v2 shows that the diversity sensitive loss is necessary for style diversity, even though the discriminator receives diverse real instances.
The visual quality of the images generated by StarGAN v2 is quite high, and the FID \cite{heusel2017gans} reaches $23.8$ on the CelebA-HQ test set \cite{karras2017progressive}. We adopted the same Residual Network-based modules from StarGAN v2 \cite{he2016deep}, but instead of using a mapping network to transfer the latent code to the style code, we use the latent code as the expression code directly.

\section{Proposed method}
We postulate that the difference between the neutral face and the emotional face for the same identity is expression. Based on that, the aim of our proposed network is to learn a unified and disentangled expression representation while excluding identity information.

\subsection{Modules}
Our network contains 3 modules: an encoder (E), a generator (G), and a discriminator (D). We use $x_e$ to represent the emotional face and $x_n$ to represent the neutral face. The expression code is a $d$-dimensional vector $c \in \mathbb{R}^d$, and we set $\vec{0}$ as the expression code for the neutral face.

\textbf{Encoder.} The encoder E takes a real or fake emotional face as input and outputs an expression code $\hat{c}=E(x_e)$ (The hat means the vector is produced by the network). The training goal of the encoder is to extract a pure expression code that does not contain any identity information.

\textbf{Generator.} The generator G takes an emotional face $x_e$ or neutral face $x_n$ and an expression code $c$ as input and output a fake face, $\hat{x_e}=G(x_n,c)$ or $\hat{x_n} =G(x_e,\vec{0})$. Note that the generator can not transfer an emotional face to another emotional face directly. This design simplifies the task of the generator since it only needs to learn two one-to-many mappings (one neutral expression to many emotional expressions and vice versa) rather than a many-to-many mapping (many emotional expressions to many emotional expressions) over each identity.

\textbf{Discriminator.} Our discriminator D is a multi-task discriminator \cite{liu2019few,mescheder2018training}. It takes a real or fake face and a face type label ($0$ for natural face; $1$ for emotional face) as input and outputs whether the input is a real image or fake one, such as $\hat{y}=D(x_e,1)$.

\subsection{Structure}

The proposed architecture is presented in \cref{fig:structure} and it consists of 2 cycles, where the encoder, generator and discriminator are shared among these two cycles:

\textbf{(a) Neutral to Emotional to Neutral cycle (N to E to N).} As \cref{fig:structure}(a) shows, this cycle transfers a neutral face to an emotional face and then transfers it back to the neutral face for the same identity. In the forward flow, a multi-dimensional expression code $\tilde{c}$ is sampled from a uniform or Gaussian distribution. The discriminator $D$ will receive real emotional faces randomly from the dataset, so the generator $G$ has to transfer the neutral face to an emotional face to compete with the discriminator, which is achieved through an adversarial loss
\begin{equation}
  \begin{split}
  \mathcal{L}_{adv\_e} = \:&\mathbb{E}_{x_e}[\log(D(x_e,1))]\:+\: \\
  & \mathbb{E}_{x_n,\tilde{c}}[\log(1-D(G(x_n,\tilde{c}),1)].
  \end{split}
  \label{eq:loss_adv_e}
\end{equation}
In the bottom backflow, similar to CycleGAN, the fake face is transferred back to the original face at the pixel level. The backflow forces the fake face to be identity-preserving by comparing it with the input face. We take the pixel level $L_1$ distance as the cycle consistency loss
\begin{equation}
  \mathcal{L}_{cyc\_n}=\mathbb{E}_{x_n,\tilde{c}}[||x_n-G(G(x_n,\tilde{c}),\vec{0})||_1].
  \label{eq:loss_cyc_n}
\end{equation}
In the upper backflow, the encoder $E$ extracts the expression code out from the fake emotional face. Similar to InfoGAN, we compare the extracted code with the random expression code to maximise the mutual information, and thus the code will correspond to the expression on the generated face, this forms the code loss
\begin{equation}
  \mathcal{L}_{code}=\mathbb{E}_{x_n,\tilde{c}}[||\tilde{c}-E(G(x_n, \tilde{c}))||_1].
  \label{eq:loss_code}
\end{equation}
Similar to StarGAN v2, we use the diversity sensitive loss \cite{mao2019mode,yang2019diversity} to let the generator $G$ generate different expressions for the same identity, which is achieved by pushing two fake emotional faces of the same person away from each other. The difference between expressions is minor compared to the difference between identities (average Euclidean distance ratio is around $1:2$), 
this further makes the diversity sensitive loss indispensable. 
\begin{equation}
  \begin{split}
  \mathcal{L}_{ds}=\:&\mathbb{E}_{x,\tilde{c_1},\tilde{c_2}}[
||G(x_n,\tilde{c_1})-G(x_n,\tilde{c_2})||_1],
  \end{split}
  \label{eq:loss_ds}
\end{equation}$\tilde{c_1}$ and $\tilde{c_2}$ are sampled independently.
The full objective of the N to E to N cycle is
\begin{equation}
  \begin{split}
  \min_{E,G}\,\max_{D}\;&\lambda_{adv\_e}\,\mathcal{L}_{adv\_e}+\lambda_{cyc\_n}\,\mathcal{L}_{cyc\_n}\\ &+\lambda_{code}\,\mathcal{L}_{code}-\lambda_{ds}\,\mathcal{L}_{ds},
  \end{split}
  \label{eq:objective_0}
\end{equation}where $\lambda_{adv\_e}$, $\lambda_{cyc\_n}$, $\lambda_{code}$ and $\lambda_{ds}$ are weights for each term. 
We observed that the generator learns to generate identities at first and then gradually learn to generate diverse expressions. We employed a dynamic code loss weight $\lambda_{code}$, which will gradually increase from $0$ to a constant. Thus the effect of the code loss in the total loss will align with the learning of the generator. In this cycle, we do not use the expression code of the real reference face but only the random code $\tilde{c}$, which is for reducing bias, since it prevents the network from paying attention to expression codes from the training samples.

\textbf{(b) Emotional to Neutral to Emotional Cycle (E to N to E).} As \cref{fig:structure}(b) shows, this cycle transfers an emotional face to a neutral face and then transfers it back for the same identity. The discriminator $D$ will receive real neutral faces, and thus it forces the generator $G$ to transfer the emotional face to a neutral face; which is implemented by an adversarial loss
\begin{equation}
  \label{eq:loss_adv_n}
  \begin{split}
  \mathcal{L}_{adv\_n} = \:&\mathbb{E}_{x_n}[\log(D(x_n,0))]\:+\: \\
  & \mathbb{E}_{x_e}[\log(1-D(G(x_e,\vec{0}),0)].
  \end{split}
\end{equation}The encoder $E$ extracts the expression code from the real face and then the generator uses it to transfer the fake face back to the emotional face, this forms a cycle loss
\begin{equation}
  \mathcal{L}_{cyc\_e}=\mathbb{E}_{x_e}[||x_e-G(G(x_e,\vec{0}),E(x_e))||_1]
  \label{eq:loss_cyc_e}
\end{equation}
The cycle loss $\mathcal{L}_{cyc\_e}$ is not only for preserving identity but also lets the encoder learn the expression of the identity. The full objective function for this cycle is
\begin{equation}
  \min_{E,G}\,\max_{D}\;\lambda_{adv\_n}\,\mathcal{L}_{adv\_n}+\lambda_{cyc\_e}\,\mathcal{L}_{cyc\_e},
  \label{eq:objective_1}
\end{equation} $\lambda_{adv\_n}$ and $\lambda_{cyc\_e}$ are weights.

In the training phase, the two cycles are combined together within each iteration. Our model needs two steps-translation to translate an emotional face into another emotional face. The first step is converting an emotional face to a neutral face and the second step is applying the expression code. We can translate the expression directly in one step by changing the structure of the network, but that may degrade the quality, as we discussed in \cref{sec:discussion}.


\section{Experiments and results}

\begin{figure*}[!ht]
  \centering
  

  
    \includegraphics[width=\textwidth]{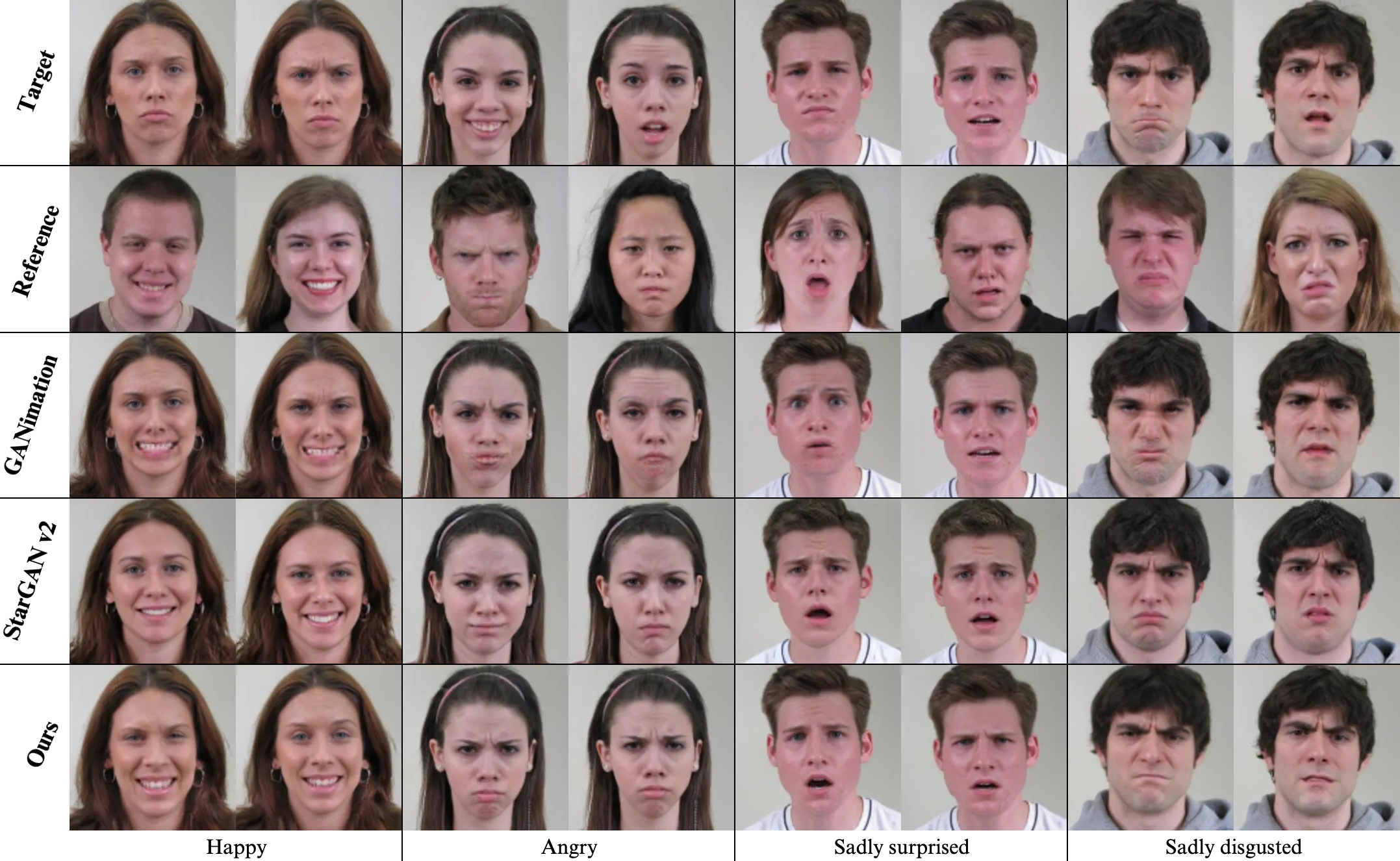}
  \caption{\textbf{Qualitative comparison} of GANimation, StarGAN v2 and our approach on the CFEE training set. The first row is real target faces. On the second row, 2 real reference faces are selected for each of the four emotions: happy, angry, sadly surprised and sadly disgusted. StarGAN v2 failed to keep identities. All 3 methods can generate differences within the same emotion, which means they can capture expression details.}
  \label{fig:QualitativeComparison}
\end{figure*}

\subsection{Datasets}
We have used two datasets to train and test the proposed model.

\textbf{CFEE:} The Compound Facial Expressions of Emotions Database (CFEE) \cite{du2014compound} is used as the main evaluation dataset in our experiments. CFEE contains 230 individuals of different races, each individual has $26$ expressions (including the neutral expression), and all faces are in a frontal pose.
We mix all emotional faces as the emotional face group and all neutral faces as the neutral face group to build our dataset. Then the training and test set are split at the identity level: $222$ individuals for the training set; $8$ individuals for the test set. The identity-level splitting is for generalization validation.

\textbf{RaFD:} Radboud Faces Database (RaFD) \cite{langner2010presentation} contains 67 individuals, each with 8 expressions. The photos are taken from 5 camera angles and each expression was shown with 3 gaze directions. We use 3 camera angles: $45^\circ$, $90^\circ$ and $135^\circ$ in our experiment.
We take neutral faces from all 3 camera angles in the frontal gaze direction as the neutral face group and the rest of the images as the emotional face group. We also split the data at the identity level: $61$ individuals for the training set; $6$ individuals for the test set.

\subsection{Training}


For both datasets, input images are resized to $128\times 128$ resolution with random cropping and horizontal flipping. The random expression code is sampled from a $32$ dimensional uniform distribution $U(-0.5,0.5)^{32}$. We used the uniform distribution instead of the Gaussian distribution to provide an equal chance to train extreme expressions, and the analysis of the components (\cref{sec:components_analysis}) showed uniform distribution works better. The code loss weight $\lambda_{code}$ is set to $0$ at step $1$ and then linearly increases to $1.0$ at step $5000$. The diversity sensitive loss ${\lambda}_{ds}$ is set to $1.0$ at step $1$, and then linearly decreases to $0$ at step $10000$. The Encoder will convert the image to grayscale before computation. More details of hyperparameters can be found at the \hyperref[code_url]{code base}. The network took 3 days to train on a single Tesla V100-16GB for 100K iterations (steps) on each dataset. We compute the Fréchet Inception Distance (FID, lower is better), \cite{heusel2017gans} on the test set after each $5000$ step. The lowest point on the CFEE test set steps $65000$, which was used in further analysis.

\subsection{Baseline Models}
We adopt GANimation and StarGAN v2 as two state-of-the-art baseline models. Both of these are pixel-level GAN models that can perform expression transfer tasks. Both models are trained with their original implementations\footnote{\url{https://github.com/albertpumarola/GANimation}}
\footnote{\url{https://github.com/clovaai/stargan-v2}}. Both models can transfer an emotional face to another emotion directly. The main drawback of these two models is that they are both supervised and require the emotion label. In our approach, we address this issue by making the model unsupervised. In addition, we can control the intensity of the generated emotion.

\textbf{GANimation} adopts a Conditional GAN (CGAN) \cite{mirza2014conditional} architecture, which takes Action Units (AUs) as the condition.
AUs describe continuous anatomical facial movements, which is a precise and meaningful expression representation. GANimation relies on third-party tools to extract AUs from the image, such as OpenFace\footnote{\url{https://github.com/TadasBaltrusaitis/OpenFace}} \cite{baltrusaitis2018openface}. 
GANimation also employs a cycle consistency loss for preserving identity. We trained GANimation on the CFEE dataset for $750$ epochs.

\textbf{StarGAN v2} learns the cross-domain style transfer in a supervised manner. It adds a mapping network to map latent codes to style codes for different domains, which is combined with a diversity sensitivity loss for generating diverse styles within the same domain. We train StarGAN v2 on the CFEE dataset by taking $26$ expression labels as $26$ domains (the training on two domains, neutral face domain and emotional face domain, failed to produce diverse expressions).


\subsection{Qualitative evaluation}
Our expression transfer results on the CFEE and RafD datasets are shown in \cref{fig:demo_matrix_1}. We observed that our method could transfer expressions over different races on the CFEE dataset and over different angles and ages on the RafD dataset.
And as shown by the child in the third row and third column of RafD, it can transfer gaze directions as well.
The result also shows deep learning-based methods can derive unknown regions, such as teeth.

\textbf{Qualitative comparison.} In \cref{fig:QualitativeComparison} a qualitative comparison of GANimation, StarGAN v2 and our approach is demonstrated.
The output of GANimation is not stable, especially when the input expression and output expression differ greatly, for example, the face on the third column shows this issue clearly. StarGAN v2 performs poorly on identity-preserving compared to our approach and although it retains the overall style (hair, skin colour, etc.), the details of the face characteristics are altered.
One possible reason for our higher identity-preserving quality is that our method focuses on converting natural faces to emotional faces and vice versa, avoiding learning the many-to-many conversion mapping done by GANimation and StarGAN v2.
In addition, StarGAN v2 consumes computations in unnecessary translations from and to the same expression. All three methods can capture diverse expression details, the differences between fake faces of the same emotion show that. StarGAN v2 employs a mapping network to map the latent code to the expression code while we take the latent code as the expression code directly, our experiment shows the mapping network can not help to produce diverse expressions in our architecture.

\subsection{Quantitative comparison}
We also conducted some quantitative analysis. We generated two fake image sets by GANimation, StarGAN v2, and our network to compute Fréchet Inception Distance (FID, lower is better) \cite{heusel2017gans} and Inception Score (IS, higher is better) \cite{barratt2018note}. The two fake image sets are: (1) fake neutral faces generated by emotional faces $\hat{x_n}=G(x_e)$; and (2) fake emotional faces generated by neutral faces and reference faces $\hat{x_e}=G(x_n,E(x_e))$. The result of the FID score comparison is shown in \cref{tab:FID_compare}, which indicates the images generated by our approach are closer to the real distribution than StarGAN v2. AUs as a precise expression representation help GANimation achieve a better FID score on fake emotional faces.
But the IS, which implies the quality of generated images, for GANimation is lower than StarGAN v2 and our approach.

\begin{table}[t]
\centering
  \begin{tabularx}{\linewidth}{|X|>{\centering\arraybackslash}p{0.1\linewidth}|>{\centering\arraybackslash}p{0.1\linewidth}|>{\centering\arraybackslash}p{0.1\linewidth}|>{\centering\arraybackslash}p{0.1\linewidth}|}
  \hline
  \multirow{2}{*}{\textbf{Method}} & \multicolumn{2}{c|}{\textbf{FID}} & \multicolumn{2}{c|}{\textbf{IS}} \\
  \cline{2-5}
  & N & E & N & E\\
  \hline
  GANimation & 51.5 & 29.1 & 1.48 & 1.49 \\
  StarGAN v2 & 55.4 & 35.4 & 1.54 & 1.54 \\
  Ours & 44.4 & 31.1 & 1.49  & 1.55 \\
  \hline
  Real images & - & - & 1.62 & 1.61 \\
  \hline
  \end{tabularx}
  \caption{\textbf{FID and IS comparison} of GANimation, StarGAN v2 and ours on the CFEE test set. The error of FID is around $\pm 0.3$, and the error of IS is around $\pm 0.03$. N: fake neutral face; E: fake emotional face generated by reference face.}
  \label{tab:FID_compare}
\end{table}

\begin{table}[t]
  \centering
  \begin{tabularx}{\linewidth}{|X|>{\centering\arraybackslash}m{0.315\linewidth}|>{\centering\arraybackslash}m{0.315\linewidth}|}
  \hline
  \textbf{Method}
  &\textbf{Identity Preserving Score}
  &\textbf{Expression Transfer Score}\\
  \hline
  GANimation & 4.62 (0.87)  & 3.56 (1.54) \\
  StarGAN v2 & 4.64 (1.06)  & 3.42 (1.62)  \\
  Ours & 4.82 (0.64) & 3.37 (1.69) \\
  \hline
  \end{tabularx}
  \caption{\textbf{Human evaluation} of GANimation, StarGAN v2 and ours on the CFEE dataset. Each score is in the range $[0,5]$. The number in the bracket is the standard deviation, which implies all models perform more stable on identity-preserving compared to expression transfer.}
  \label{tab:Human_evaluation}
\end{table}

\begin{figure*}[!ht]
  \centering
  \includegraphics[width=\textwidth]{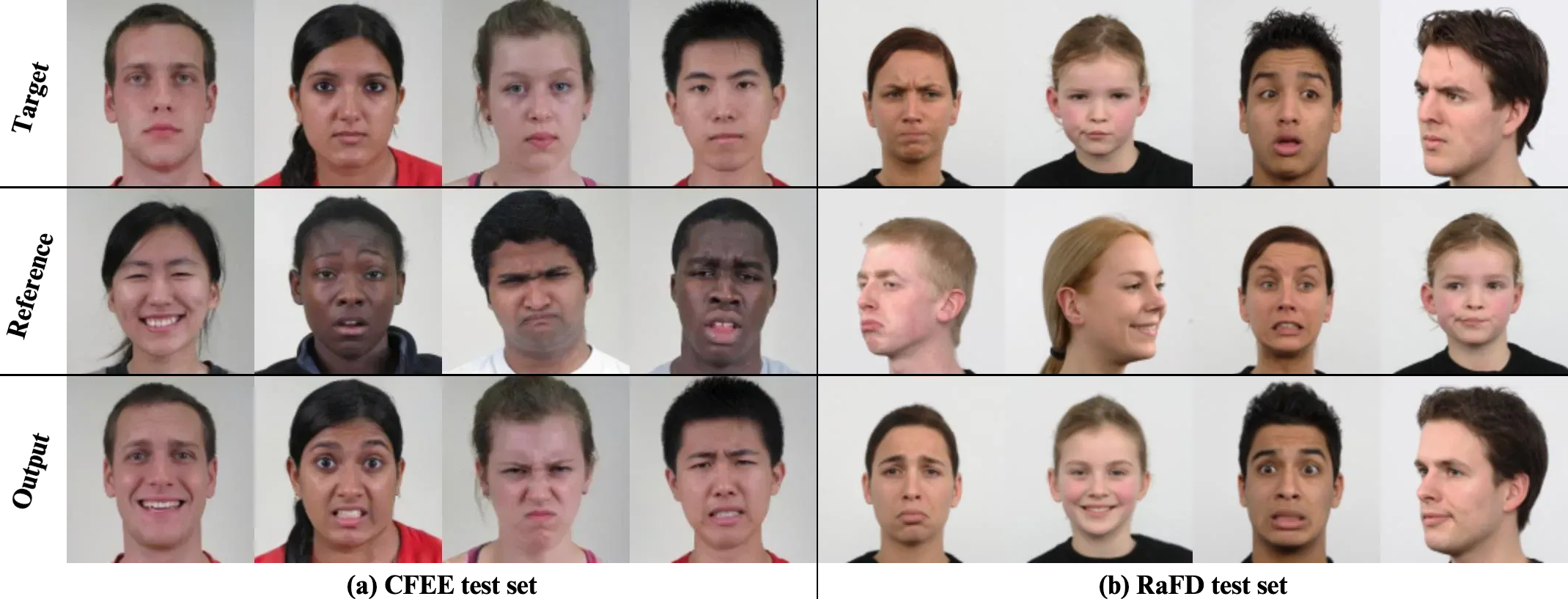}
  \caption{\textbf{Qualitative evaluation of generalization} on the CFEE and RaFD test set. The first row is real target faces, the second row is real reference faces, and the last row is our outputs. None of the identities appeared during the training phase. It shows our network is able to generalize to unknown identities.}
  \label{fig:QualitativeGeneralization}
\end{figure*}

\textbf{Human evaluation.}
We further evaluated the generated samples based on the CFEE dataset by 25 workers on Amazon Mechanical Turk (AMT). We focused on two aspects which are identity preservation quality and expression transfer quality.
We asked the workers to evaluate each sample using a 5-point Likert-scales from "Definitely not" to "Definitely yes".
We randomly selected 3 images for each of the 25 emotions as reference faces and randomly selected one target face for each reference face. Then, we used GANimation, StarGAN v2 and our model to generate samples for each reference and target pair.
The total number of generated samples is 225. We shuffled these samples along with 10 extra quality control samples. The purpose of the quality control questions is to assess the workers' attention to the task, and we filtered out 6 workers whose response rates to the quality control samples was very low.
For each question, we removed outliers using a statistical outlier analysis by removing samples with a z-score above $3$ or below $-3$ and then averaged the remaining scores as the final score. This ensures that those evaluations that are far away from the rest are excluded.
As shown in \cref{tab:Human_evaluation}, the results are consistent with our observations. Our model achieved the highest identity preservation score among the three models, the expression transfer score is close to the supervised learning model StarGAN v2. The AUs-guided model GANimation is more accurate in expression transfer but over-distorted in some samples.

\subsection{Components analysis}
\label{sec:components_analysis}
To demonstrate the effect of different components of the model, we cumulatively change a component in our final model to analyse the impact on the generated results.
The FID and IS are computed after each modification as reported in \cref{tab:Components_analysis}. The changed components are listed below:
\begin{itemize}
  \item "dynamic loss weights". The dynamic loss weights including: $\lambda_{code}$ increase from $0$ at step $1$ to $1.0$ at step $5000$; $\lambda_{cyc\_e}$ increase from $1.0$ at step $1$ to $1.5$ at step $10000$. In the test, we set both weights to a constant $1$.
  \item "grayscale encoder". The encoder transfer images to grayscale before computing, which reduces computation and lets the encoder focus on useful information. In the test, we changed it to a color encoder.
  \item "uniform distribution". Sampling expression codes from a uniform distribution, which can provide an equal opportunity to train expressions of different strengths. In the test, we switch it to a Gaussian distribution.
\end{itemize}
FIDs show that each component makes generated images closer to the true distribution on the test set, especially the grayscale encoder component.

\begin{table}[t]
\centering
  \begin{tabularx}{\linewidth}{|X|l|l|}
  \hline
  \textbf{Component} & 
  \multicolumn{1}{c|}{\textbf{FID}} &
  \multicolumn{1}{c|}{\textbf{IS}}\\
 \hline
  final model & 32.4 & 1.53 \\
  - dynamic loss weight & 33.5 (+1.1) & 1.54 (+0.01) \\
  - grayscale encoder & 37.3 (+3.8) & 1.57 (+0.03) \\
  - uniform distribution & 38.0 (+0.7) & 1.49 (-0.08) \\
  \hline
  \end{tabularx}
  \caption{\textbf{Components analysis.} Both FID and IS are computed after each modification.}
  \label{tab:Components_analysis}
\end{table}

\subsection{Generalization of unknown identities}  
Some samples of the generated fake faces from the CFEE and RaFD test sets are provided in \cref{fig:QualitativeGeneralization}. None of these identities was presented during the training. We observed that our network could generalize to unknown identities for expression extraction and generation. However, the examples generated from the test set did not achieve the same level of identity-preserving as in the training set, especially on the RaFD dataset. One reason could be the CFEE training set contains only 214 individuals, and the RaFD training set has only 61 individuals, which is minuscule compared to the whole population.



\subsection{Expression continuity}
\begin{figure}[t]
  \centering
  \includegraphics[width=\linewidth]{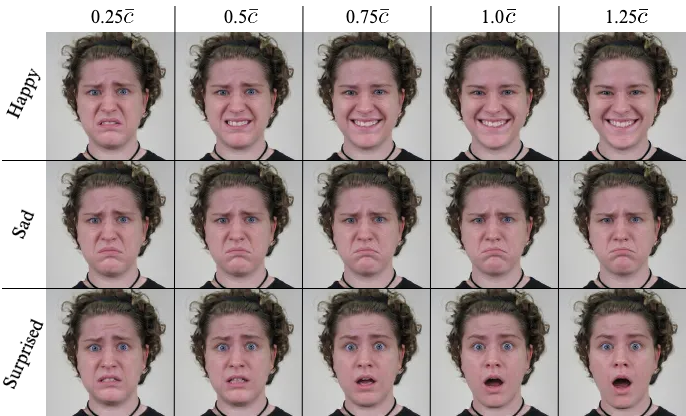}
  \caption{\textbf{Continuity analysis} on the CFEE dataset. Each row is a different emotion. The expression code increases linearly from left to right, and that results in controlling the strength of the deformation.}
  \label{fig:code_face_CFEE}
\end{figure}

For each emotion, we took the mean of all extracted expression codes on the training set as the standard expression code $\bar{c}$. Then, we scaled it in or out (e.g. $0.5\cdot\bar{c}$) and then applied it on a neutral face. As \cref{fig:code_face_CFEE} shows, there is a continuous linear relationship between the expression code values and the resulting deformations.

\section{Discussion and Limitation}
\label{sec:discussion}
The combination of random expression codes and diversity sensitive loss facilitates the network to explore the entire instance space in order to produce diverse expressions.
Explicitly converting the many-to-many mapping problem to two one-to-many mapping problems helps reduce the learning burden.
The network learns the differences between two groups, which could include details such as posture, gaze direction, tears, etc.


The nature of our model is to learn the transfer of deformation rather than the transfer of emotion. We speculate this is one of the reasons that our model achieves a lower expression transfer score on some of the faces, since two faces being close on deformation does not necessarily mean they are expressing the same emotion. We used very neat training sets to train our model, for the  CFEE training set, each identity in the neutral face group has 25 corresponding expressions in the emotional face group. If the identities appear in the neutral face group then do not appear in the emotional face group, the training will fail in identity-preserving.
To train a general and applicable network, there is a need for a very diverse dataset for capturing all poses, races, ages, lighting conditions and backgrounds etc., which could be a difficult task; however, this is a general issue with all expression generation models.

\section{Ethics}
The proposed method could be used to generate fake expressions of individuals. The application's regulations have been discussed at \cite{meskys2020regulating}.
\section{Conclusion}
In this paper, we propose a novel GAN-based network to transfer expressions between human faces, which aims to address two major problems: unsupervised learning of expression transfer and generating continuous and diverse expressions.
We combine CycleGAN \cite{zhu2017unpaired} and InfoGAN \cite{chen2016infogan} in a novel way to solve the identity-preserving problem and the unsupervised learning problem. The experiment shows our network can be trained without emotion labels and can transfer continuous and detailed expressions on CFEE and RaFD datasets. The evaluation shows the generated expressions by our approach are closer to the real distribution and have a higher identity preservation quality compared to GANimation and StarGAN v2, and yet its expression transfer quality is comparable to them.
For further research, removing the requirement of neutral face labels and testing the network on more diverse situations, such as videos can be explored.\\
\\

{\small
\bibliographystyle{ieee_fullname}
\bibliography{egbib}
}

\end{document}